\documentclass[preprint,12pt]{elsarticle}

\usepackage{amssymb}
\usepackage{amsmath}
\usepackage{hyperref}
\usepackage{url}
\usepackage{multirow} 
\usepackage{array}    
\usepackage{graphicx}
\usepackage{subcaption}
\usepackage{algorithm}
\usepackage{xcolor}
\usepackage{algorithmic}
\usepackage{bm}

\journal{Neurocomputing}

\begin{document}

\begin{frontmatter}



\title{Robust Emotion Recognition via Bi-Level Self-Supervised Continual Learning}

\author[label1]{Adnan Ahmad \corref{cor1}}
\author[label1]{Bahareh Nakisa}
\author[label2]{Mohammad Naim Rastgoo}

\cortext[cor1]{Corresponding author: Adnan Ahmad (email: adnan.a@deakin.edu.au)}
\affiliation[label1]{organization={School of Information Technology},
					addressline={Deakin University},
					city={Geelong},
					country={Australia}
}

\affiliation[label2]{organization={Faculty of Information Technology},
            addressline={Monash University},
            country={Australia}}


\begin{abstract}
   Emotion recognition through physiological signals such as electroencephalogram (EEG) has become an essential aspect of affective computing and provides an objective way to capture human emotions. However, physiological data characterized by cross-subject variability and noisy labels hinder the performance of emotion recognition models. Existing domain adaptation and continual learning methods struggle to address these issues, especially under realistic conditions where data is continuously streamed and unlabeled. To overcome these limitations, we propose a novel bi-level self-supervised continual learning framework, SSOCL, based on a dynamic memory buffer. This bi-level architecture iteratively refines the dynamic buffer and pseudo-label assignments to effectively retain representative samples, enabling generalization from continuous, unlabeled physiological data streams for emotion recognition. Then assigned pseudo-labels are subsequently leveraged for accurate emotion prediction. Key components of the framework, including a fast adaptation module and clusters mapping module, enable robust learning and effective handling of evolving data streams. 
    Experimental validation on two mainstream EEG tasks demonstrates the framework’s ability to adapt to continuous data streams while maintaining strong generalization across subjects, outperforming existing approaches.  
\end{abstract}




\end{frontmatter}


\section{Introduction}

Emotion recognition through physiological data has become a cornerstone of affective computing due to its ability to objectively capture human emotions \cite{Nakisa2018LongST,Hay2012EmotionRI,Nakisa2018EvolutionaryCA}. Unlike non-physiological signals such as text, video, and audio, which are often influenced by subjective factors and do not reliably reflect the true emotional state, physiological signals such as the electroencephalogram (EEG) provide a more accurate and unbiased representation of emotions \cite{Deng2018SemisupervisedAF}. These signals are increasingly being used to develop intelligent systems that adapt to human stress and cognitive load in dynamic, high- stakes environments such as air traffic management \cite{Aric2016APB,Aric2016AdaptiveAT}, transportation \cite{Rastgoo2019AutomaticDS,Mou2023DriverER}, search and rescue operations \cite{lim2021cognitive}, and entertainment \cite{Hafeez2021EEGIG}.

Classical machine learning (ML) approaches to emotion recognition rely heavily on large, annotated datasets. Publicly available datasets are typically collected under controlled conditions using specialized emotion elicitation techniques, which limits their generalizability to different real-world scenarios \cite{khare2024emotion}. In addition, physiological signals vary significantly across individuals, even when experiencing the same emotional state, leading to poor performance in cross-subject emotion recognition \cite{Zhang2018IndividualER,Liu2024VBHGNNVB}. While domain adaptation methods \cite{Zhao2021PlugandPlayDA,Li2020DomainAF} have shown promise in addressing this problem, they often require extensive data collection from new subjects to customize models, making them impractical for real-world applications. Some researchers have explored Online Continual Learning (OCL) techniques to address these challenges by training emotion recognition models on continuous data streams \cite{Duan2024RetainAA,Duan2024OnlineCD}. However, existing OCL approaches assume that multiple emotion classes are consistently present in each data stream, which may not accurately reflect real-world conditions. In practice, a single dominant emotion is typically elicited at any given time, with occasional overlaps occurring during transitions between emotional states. This discrepancy can impact the effectiveness of such approaches in real-world applications. Moreover, these methods are based on supervised learning and rely on subjective labels, which are often noisy \cite{Jiang2024REmoNetRE} and may not correspond well with actual emotional states as they are derived from subjective assessments during data collection. \textcolor{black}{Furthermore, conventional self-supervised learning approaches in EEG often rely on data augmentation (DA) techniques that distort the temporal characteristics of EEG signals, such as random cropping or scaling. These augmentations can degrade signal fidelity, leading to suboptimal representations in low signal-to-noise environments  \cite{goldenholz2009mapping}.}

Unsupervised Continual Learning (UCL) has been applied in various domains to extract knowledge from raw data \cite{Jiang2016VariationalDE}. However, deploying existing UCL algorithms in real-world scenarios to learn over time remains challenging for several reasons, one of which is their assumption of distinct class boundaries \cite{Tiezzi2022StochasticCO}. For instance, when class boundaries are clearly defined and known, the learning algorithm can expand the network or create new memory buffers when a class shift is detected. In contrast, physiological signal streams have no clear boundaries, as emotions gradually transition, resulting in blurred or overlapping boundaries. Developing generalizable emotion recognition models that are able to cope with cross-subject variability in such complex scenarios remains a significant challenge. This paper addresses these challenges by focusing on  non-independent and identically distributed (non-iid) input data streams, where samples occur only once, class labels are unavailable, and emotions transition without clear boundaries.


This paper presents SSOCL, a novel self-supervised OCL framework for emotion recognition from sequential physiological data streams. SSOCL uses a bi-level architecture that dynamically maintains a compact memory buffer in which only the most representative samples are stored together with their pseudo-labels. In the inner loop, SSOCL adapts to incoming data using a self-supervised learning module combined with a cluster assignment strategy to assign pseudo-labels to the samples in the current batch. \textcolor{black} {The self-supervised learning module exploits temporal dependency in EEG streams by predicting future embeddings from current representations, enabling robust feature learning without relying on naive augmentations that may distort the temporal structure of the signal.} A cluster mapping module ensures consistency by aligning newly formed clusters with pseudo-labels from previous data streams. In the outer loop, a memory enhancement module selectively updates the buffer to preserve the most important pseudo-labeled samples, ensuring efficient use of the limited memory capacity. The memory buffer is used for replay alongside the input data stream to train the emotion recognition model, enabling strong generalization across evolving data streams.
Our key contributions are summarized as follows:
\begin{itemize}
    \item We introduce a Self-Supervised Online Continual Learning (SSOCL) framework that enables the model to adapt to incoming data streams without relying on labeled data. This approach addresses the challenges posed by noisy and subjective labels, allowing the model to learn directly from raw EEG signals in practical, real-time scenarios. 
    \item SSOCL consists of a bi-level architecture that dynamically manages a compact memory buffer in which only the most representative samples are stored together with the corresponding pseudo-labels. To ensure consistency across evolving data streams, we integrate a cluster assignment module that maps newly formed clusters with pseudo-labels from previous data. In addition, a memory enhancement module selectively updates the buffer to retain the most generalized samples.
    \item \textcolor{black}{We design a novel self-supervised learning module that captures temporal dependencies in EEG streams without relying on traditional data augmentation. Instead of augmenting samples, our approach learns by predicting future embeddings within a batch using a predictor network and contrastive loss. This design is tailored for low signal-to-noise EEG data, enabling more robust and meaningful representation learning in streaming, label-free scenarios.}
    \item Our framework is validated on two mainstream EEG tasks, demonstrating robust adaptation to continual data streams and superior generalization across all seen data.
\end{itemize}
\begin{figure*}[t]
  \centering
  \includegraphics[width=\textwidth]{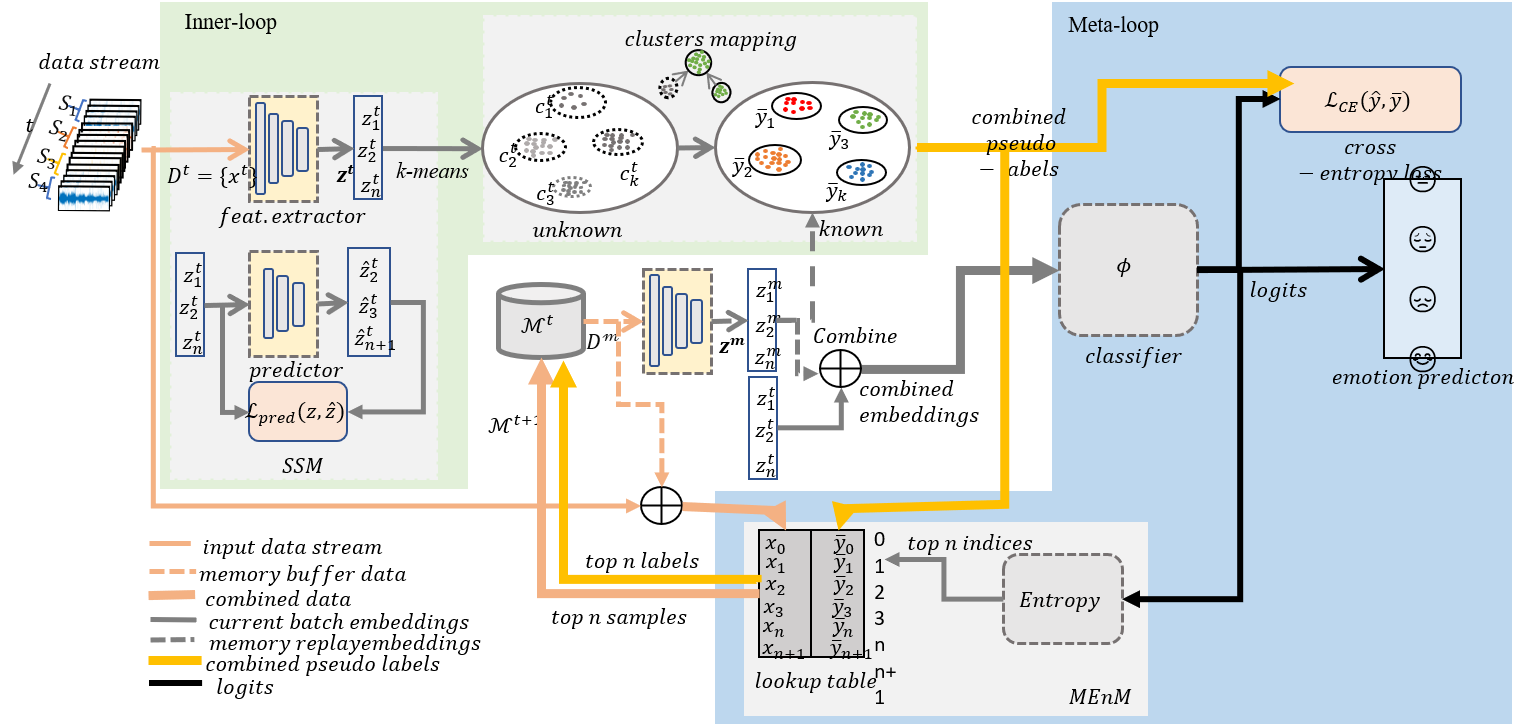}
  \caption{SSOCL framework: \textbf{Inner-loop:} In a time step $t$, the feature extractor adapts to the current data $D^t = {x^t}$ in a self-supervised manner by using a self supervised module (SSM), which consists of a predictor network that takes current embeddings $z_1^t$ and predicts future embeddings $\hat{z}_2^t$. After adaptation, k-means clustering partitions $\bm{Z}^t$ into $k$ clusters. To assign the pseudo labels, the cluster mapping module maps the newly formed clusters to the existing data stored in a memory buffer. \textbf{Meta-loop:} The current embeddings and the memory data embeddings are combined for emotion prediction. Finally, a memory enhancement module (MEnM) refines the samples for replay in the next time step $t+1$.}

  \label{fig:framework}
\end{figure*}

\section{Related Work}\label{sec:related_work}
\subsection{Cross-subject Emotion Recognition}
Cross-subject variations in EEG signals present significant challenges for cross-subject emotion recognition \cite{Jia2022HybridSN}. Several approaches have been proposed to enhance the generalizability of models. \cite{Li2020DomainAF} focuses on Domain Adaptation (DA) minimizing classification errors for the source domain while aligning source and target data in latent spaces. \cite{Li2020MultisourceTL} utilizes source selection and style transformation mapping to reduce domain discrepancies between target and source data, facilitating the extraction of shared features from EEG signals.  \cite{Zhao2021PlugandPlayDA} introduces a DA approach that customizes the model using EEG from target subjects, improving both specificity and generalization. \cite{Li2020ANT} presents a Transferable Attention Neural Network (TANN), which employs local and global attention mechanisms to dynamically emphasize discriminative EEG features for emotion recognition. These methods address the complexities of cross-subject emotion recognition, contributing to more robust and adaptable models. However, these methods are primarily limited to offline learning and require stationary and annotated target data, which limits their adaptability to dynamic real-time scenarios.
\subsection{Self-supervised Learning}
Self-supervised learning has also shown promise in learning generalized and robust representations from unlabeled data. 
Techniques such as contrastive learning \cite{Sarkar2020SelfSupervisedER,DBLP:conf/icml/ChenK0H20,Wang2023SelfSupervisedEE}, and generative adversarial networks (GANs) \cite{Zhang2021GANSERAS} have been explored to train models effectively. For instance, \cite{Sarkar2020SelfSupervisedER} used different transformations on physiological signals to generate pretext tasks, followed by training a Convolutional Neural Network (CNN) to recognize signal variations. \cite{Wang2023SelfSupervisedEE} applied similar self-supervised methods for EEG emotion recognition, using pretexts to generate labels from unlabeled EEG data. 
Transformer-based self-supervised learning has been adapted for multimodal emotion recognition to improve generalization across tasks\cite{VazquezRodriguez2022TransformerBasedSL,Wu2023TransformerBasedSM}. However, these works require labeled datasets to fine-tune models on downstream tasks. Furthermore, CLUDA \cite{Ozyurt2022ContrastiveLF} is a contrastive learning-based DA approach for time series data that aligns contextual representations between source and target domains through adversarial training. However, this method also requires static data, limiting their applicability in continuous, real-time scenarios.
\subsection{Online Continual Learning}
Very few studies have been conducted in the literature on OCL, particularly in the context of EEG data. Cao et al. \cite{Cao2023LURAO} proposes a model that selectively forgets unimportant weights and retrains on continuous data streams. However, this approach struggles with memory limitations due to storing all previous data, leading to potential bottlenecks. AMBM is a meta-learning approach introduced to address catastrophic forgetting in OCL \cite{Duan2024RetainAA}. It employs a bi-level mutual information maximization process, using a memory buffer to retain significant features for fast adaptation. A recent study proposes a method that balances data informativeness and storage across different subjects \cite{Duan2024OnlineCD}. This approach effectively tackles data imbalance in memory to reduce memory usage. Another study \cite{Wang2022ImprovingTC} addresses the challenge of maintaining a consistent memory buffer that captures generalized representations. However, these approaches are categorized into supervised learning and heavily rely on labeled data. A recent study, SCALE \cite{Yu2022SCALEOS}, addresses a practical scenario similar to ours by using contrastive losses and an efficient memory buffer for self-supervised OCL. However, the performance of this method may be limited when applied to EEG data due to its heavy reliance on data augmentation (DA) techniques. Since EEG signals have a low signal-to-noise ratio, DA could introduce additional noise and potentially affect temporal information \cite{goldenholz2009mapping}.
\section{Preliminaries}
Consider a neural network $\mathcal{F}$ that is pre-trained on a source dataset $D = \{(x_i, y_i)\}_{i=1}^{N}$ of size $N$, where $\bm{x}_i \in \mathcal{X}$ denotes a physiological signal segment in the input feature space $\mathcal{X}$, and $y_i \in \mathcal{Y}$ is the corresponding label. The model $\mathcal{F}$ consists of a feature extractor $f$ and a classifier $\phi$, and is trained to predict $K$ different emotions. We aim to adapt the model $\mathcal{F}$ in OCL where a set of subjects $\mathcal{S} = \{\mathcal{S}_1, \mathcal{S}_2, \dots, \mathcal{S}_s\}$ arrives sequentially. This setup is characterized by both class-incremental and subject-incremental learning. 
A subject $s \in \mathcal{S}$ generates unlabeled data $D_s^t = \{\bm{x}_j\}_{j=1}^{B}$ comprising $B$ data points at time $t$. At any point in time, $D_s^t$ can be viewed as the dataset $D^t$ since subjects are arriving sequentially. The union of the feature spaces of the sequentially arriving data, $\bigcup_{s \in \mathcal{S}} \mathcal{X}_s$, forms a target feature space $\mathcal{X}_{\mathcal{T}}$.
The goal is to adapt the pretrained neural network $\mathcal{F}$ to optimally generalize to the target feature space $\mathcal{X}_{\mathcal{T}}$, with the model being capable of handling the continuously evolving nature of the feature and label spaces as new subjects arrive.

Since only the feature space of the current subject, $\mathcal{X}^t \in \mathcal{X}_{\mathcal{T}}$, is known at time $t$, there is a significant risk of forgetting previous feature spaces, which can lead to a loss of generalizability of the model $\mathcal{F}$. A common approach to mitigate this issue is to maintain a memory replay buffer $\mathcal{M}$, which stores a subset of data from previously encountered feature spaces. This allows the model to learn from past data streams and helps prevent forgetting. Previous works \cite{Duan2024OnlineCD} have leveraged label information to store the most representative data in the memory buffer. However, this approach is not applicable in practice because the target label space $\mathcal{Y}_{\mathcal{T}}$ needs to be inferred from the feature space $\mathcal{X}_{\mathcal{T}}$.
In addition, due to inter-subject variability, it is challenging to maintain a representative set of data from all classes that remains generalizable across subjects. Our work aims to address this challenge by maintaining a memory buffer that stores the most representative samples from the global feature space $\mathcal{X}$, while ensuring distinct boundaries between classes. This strategy allows the model to retain important knowledge from previous unlabeled data streams while adapting to new subjects.

\section{Proposed Methodology}
The proposed framework, SSOCL, is illustrated in Figure~\ref{fig:framework}. SSOCL operates within a bi-level structure: the inner loop focuses on fast adaptation to continuous unlabeled data streams and assigning pseudo-labels to each data point in the current batch. The meta-loop evolves a memory buffer that stores the most representative samples of past data streams with their pseudo-labels while simultaneously training the model $\mathcal{F}$. 

SSOCL consists of three key components designed to handle the dynamic nature of the memory over time. First, a self-supervised learning module adapts a copy of the feature extractor to the current data stream using a future predictive loss. Next, k-means clustering is applied to partition the current batch of data $D^t$ into $k$ different clusters. A cluster mapping module then maps the newly formed clusters to the existing memory data and assigns appropriate pseudo-labels. Finally, a memory enhancement module updates the memory with the most representative samples. The model is then trained in the meta-loop using cross-entropy loss.

\subsection{Inner Objective}
This section explains the components involved in the inner objective, which aims to assign pseudo-labels to the input data stream. Two key components are involved in this process. First, a self-supervised learning module is used for fast adaptation to the current mini-batch. Then, a cluster mapping module is applied to map the clusters obtained by k-means to the existing data in memory.
\subsubsection{Self-supervised learning module} \label{sec:sslm}
As previously discussed in section \ref{sec:related_work}, various DA techniques have been developed to capture data-specific representations in self-supervised learning by imposing contrastive loss objectives to distinguish between positive and negative augmented samples. However, it is important to note that EEG signals are characterized by a low signal-to-noise ratio \cite{goldenholz2009mapping}. Data augmentation techniques such as random cropping or scaling can compromise temporal relationships, introducing additional noise. This can result in feature embeddings that fail to accurately represent meaningful temporal information. Therefore, it is important to incorporate alternative approaches to DA that go beyond traditional DA methods and focus on utilizing the temporal dependencies within the data stream.

To effectively leverage the temporal relationship between consecutive samples within a batch, we employ a predictor network $h$ that predicts the subsequent embedding $\hat{\bm{z}}_{n+1}^t$ given the current sample $\bm{z}_n^t = f(x_n^t)$. The predictor network operates by mapping the current embedding $\bm{z}_h^t$ to the next embedding $\hat{\bm{z}}_{n+1}^t$, i.e., $\hat{\bm{z}}_{n+1}^t = h(\bm{z}_n^t)$. A contrastive loss is then applied by treating $\hat{\bm{z}}_{n+1}^t$ as a positive sample for $\bm{z}_n^t$, while all other predictions $\{\hat{\bm{z}}_{0}^t, \dots, \hat{\bm{z}}_{n-1}^t,\hat{\bm{z}}_{n+2}^t, \hat{\bm{z}}_{n+3}^t, \dots, \hat{\bm{z}}_{n+B}^t\}$ (where \(B\) is the batch size) are regarded as negative samples.

\begin{equation}\label{eq:self_loss}
\mathcal{L}_{p} = -\log \frac{\exp(\hat{\bm{z}}_{n+1}^t \cdot \bm{z}_{n+1}^t / \tau)}{\sum_{j=1}^{B} \exp(\hat{\bm{z}}_{n+1}^t \cdot \hat{\bm{z}}_j^t / \tau)}
\end{equation}

where $(\cdot)$ denotes the dot product between the predicted embedding $\hat{\bm{z}}_{n+1}^t$ and the actual embedding $\bm{z}_{n+1}^t$, and $\tau$ is the temperature parameter. This loss function maximizes the similarity between the positive pair $(\bm{z}_{n+1}^t, \hat{\bm{z}}_{n+1}^t)$ while minimizing the similarity between the positive pair and all other negative pairs $(\hat{\bm{z}}_j^t, j \neq \{n+1\})$. After learning representations on the current batch of data stream $D^t$, the model $\mathcal{F}^{'}$ effectively fits $D^t$, producing embeddings $\bm{z}^t$ that better capture meaningful patterns. These improved embeddings can be exploited to identify inter-class patterns within the given batch.

It is important to emphasize that the EEG data is streamed in a class-incremental way, where small temporal segments of a specific length are treated as individual samples. Therefore, the number of distinct classes in a given batch $D^t$ at time $t$ is unknown. We perform class exploration to find inter-class patterns in each incoming batch. Specifically, we use the $k$-means clustering algorithm to explore the embedding space for potential class structures.  Given the unknown number of classes in $D^t$, we initialize $k$-means with $K$, which corresponds to the number of classes in the source dataset. This initialization partitions the embedding space $\bm{z}^t$ into a set of $K$ clusters, denoted as $\bm{\mathcal{C}}^t = \{\bm{\mathcal{C}}_1^t, \bm{\mathcal{C}}_2^t, \dots, \bm{\mathcal{C}}_K^t\}$, where each cluster $\bm{\mathcal{C}}_i^t$ represents a group of embeddings within the batch $D^t$. This initialization with $K$ serves as a starting point to explore the potential class structure in the current batch and assign pseudo-labels to each sample in $D^t$.

However, initializing $k$-means with $K$ may not always yield reliable pseudo-labels. For instance, if $D^t$ contains data from a single class, initializing with $K$ will create unnecessary partitions, with only one cluster representing the true class. However, in section \ref{sec:mem_enh}, we will discuss how the continuous evolution of the memory replay buffer will naturally create distinct boundaries between data samples, allowing the model to store more reliable pseudo-labeled data while discarding noisy labeled samples.

\subsubsection{Clusters mapping module}
Consider the initial time step $t_0$, when the first batch of data $D^{t_0}$ is processed and partitioned into $K$ distinct clusters. Each cluster is assigned a unique pseudo-label based on its cluster number, and the batch $D^{t_0} = \{x^{t_0}, \bar{y}^{t_0}\}$ is stored directly in the memory buffer $\mathcal{M}$, since it is initially empty. However, subsequent batches $D^t$ cannot be directly stored in $\mathcal{M}$ due to two main challenges. First, the $k$-means algorithm partitions $D^t$ into $K$ clusters, which provide cluster information but do not yield definitive pseudo-labels. Before merging $D^t$ into $\mathcal{M}$, it is necessary to map the clusters $\bm{\mathcal{C}}^t = \{\bm{\mathcal{C}}_1^t, \bm{\mathcal{C}}_2^t, \dots, \bm{\mathcal{C}}_K^t\}$ to the existing samples in the memory $D^m = \{x^m, \bar{y}^m\}$, ensuring consistency in the pseudo-labels. Additionally, since $\mathcal{M}$ has a fixed size, a sample replacement strategy is required when the memory buffer reaches its capacity. This strategy should prioritize retaining the most informative and reliable samples while discarding less representative or noisy labeled samples. To this end, the centroids of each cluster of the memory buffer $\mathcal{M}$ are computed based on the data stored in memory:
\begin{equation}\label{eq:centmem}
\mathbf{c}_k^m = \frac{1}{|\mathcal{D}_{k}^m|} \sum_{\mathbf{x}^m \in \mathcal{D}_k^m} f(x^m)
\end{equation}
where $\mathcal{D}_k^m$ represents the set of embeddings in memory corresponding to pseudo-label $k$. Similarly, centroids for clusters in $D^t$ are computed as:
\begin{equation}\label{eq:centclus}
\mathbf{c}_k^t = \frac{1}{|\mathcal{D}_{k}^{t}|} \sum_{\mathbf{x}^t \in \mathcal{D}_{k}^{t}} f(x^t)
\end{equation}
The clusters are mapped by comparing the cosine distance between centroids:
\begin{equation}\label{eq:cosine}
\mathcal{A}(\mathbf{c}_i^t, \mathbf{c}_j^m) = \arg\min_{i,j} \left(1 - \frac{\mathbf{c}_i^t \cdot \mathbf{c}_j^m}{\|\mathbf{c}_i^t\| \|\mathbf{c}_j^m\|}\right)
\end{equation}
Centroids with high cosine similarity are likely to belong to the same class and should be merged together, with the same pseudo-label being assigned to both. Once the clusters are mapped, the current batch $D^t = \{x^t, \bar{y}^t\}$ is merged with the memory samples $D^m = \{x^m, \bar{y}^m\}$ to form a combined set:
\begin{equation} \label{eq:combined_data}
    D = D^t \cup D^m = \{x, \bar{y}\}
\end{equation}

This finalizes the inner-loop objective, and the combined set $D$ is further utilized in the meta-loop for training the model $\mathcal{F}$. A subset of $D$ is then selected to update the memory buffer, transitioning from $\mathcal{M}^t$ to $\mathcal{M}^{t+1}$.

\subsection{Meta-Loop}

In the meta-loop, while training the model $\mathcal{F}$ on the pseudo-labeled data $D$, we incorporate a memory update mechanism that selects a representative subset of data from $D$ to ensure it reflects previously seen feature spaces. To achieve this, we first introduce the memory enhancement module.

\subsubsection{Memory Enhancement Module}\label{sec:mem_enh}
An effective sample selection strategy in must account not only for the diversity of past experiences but also for the reliability and discriminative power of each data point \(x \in D\) in enhancing the generalization capacity of the evolving model \(\mathcal{F}\). Our sample selection mechanism aims to retain most representative, low-noise samples from the streaming data to populate the fixed-size memory buffer. Inspired by ODIN \cite{liang2018enhancing}, a method originally designed for detecting out-of-distribution (OOD) samples via calibrated confidence scores, we adapt its core principles to distinguish between clean and noisy pseudo-labeled samples in our self-supervised setting. Specifically, we incorporate a temperature-scaled softmax function to calibrate the model's output distribution and amplify uncertainty estimation. For each incoming sample \(x\), we compute the temperature-adjusted softmax probabilities:
\begin{equation}
S_i(x; T) = \frac{\exp(f_i(x) / T)}{\sum_j \exp(f_j(x) / T)}
\end{equation}
where \(S_i(x; T)\) denotes the probability of class \(i\) after applying a temperature parameter \(T > 1\), and \(f_i(x)\) is the logit output of the model \(\mathcal{F}\). Increasing the temperature produces a softer probability distribution, which effectively magnifies the differences between high- and low-confidence predictions.

To quantify uncertainty, we calculate the entropy of the softened prediction:
\begin{equation}\label{eq:entro}
H(x; T) = - \sum_i S_i(x; T) \log S_i(x; T)
\end{equation}

High-entropy samples are indicative of ambiguous or poorly separated feature representations, often resulting from overlapping emotional states or noisy pseudo-labels. In contrast, low-entropy samples correspond to well-separated, high-confidence predictions that are more informative and reliable for memory retention. Our memory enhancement module employs an entropy-based filtering mechanism that systematically excludes high-entropy samples while prioritizing the retention of low-entropy instances. This selective update strategy ensures that the memory buffer contains representative and confidently classified samples that best capture the underlying structure of the data distribution seen so far.

Moreover, the use of temperature scaling not only sharpens the distinction between clean and noisy samples but also promotes the maintenance of well-defined class boundaries. This is particularly important in emotion recognition from EEG, where inter-class similarity is high, and pseudo-label noise is prevalent. By selectively excluding high-uncertainty samples, our method minimizes memory contamination, preserves salient class-discriminative features across data streams, and thereby enhances resistance to catastrophic forgetting while promoting robust continual adaptation.

\subsubsection{Supervised Training with Memory Data} \label{sec:supervised_memory}

After the memory buffer \( \mathcal{M} \) has been selectively populated with reliable, low-entropy pseudo-labeled samples, we leverage it during the meta-phase for supervised training. The combined dataset, comprised of the current stream data and memory samples, is constructed as per Equation~\ref{eq:combined_data}. This fusion ensures the model retains relevant past knowledge while adapting to the distributional characteristics of the incoming data stream.

To perform supervised learning, a mini-batch \( \mathcal{B} \) is sampled from the combined dataset. Each data point \( (x, \bar{y}) \in \mathcal{B} \) consists of an input sample \( x \) and its corresponding pseudo-label \( \bar{y} \). The model predictions \( \hat{y} \) are obtained by passing \( x \) through the network \( \mathcal{F} \). We compute the standard cross-entropy loss:

\begin{equation} \label{eq:celoss}
    \mathcal{L}_{CE} = - \frac{1}{|\mathcal{B}|} \sum_{(x,\bar{y})\in \mathcal{B}} \bar{y} \log \hat{y}
\end{equation}

Here, \( \mathcal{L}_{CE} \) quantifies the divergence between the model's predicted class probabilities and the target pseudo-labels. This supervised training step facilitates both short-term adaptability to current data and long-term retention of essential prior knowledge, enabling robust performance in dynamic, non-stationary environments typical of real-world streaming data. The overall framework is summarized in Algorithm \ref{alg:ssocl}.

\begin{algorithm}[t]
\caption{\textcolor{black}{Self-Supervised Continual Learning (SSOCL)}}
\label{alg:ssocl}
\begin{algorithmic}[1]
\STATE \textbf{Input:} Data streams $\{D^t\}_{t=1}^{T}$, source model $\mathcal{F}$, $\mathcal{M} \gets \emptyset$ \COMMENT{Empty memory buffer}
\STATE \textbf{Output:} Updated model $\mathcal{F}$, memory buffer $\mathcal{M}$.
\STATE \textbf{Initialize:}
\FOR{each incoming batch $D^t$}
    \STATE Extract embeddings $\bm{z}_n^t = f(x_n^t)$.
    \STATE Compute contrastive loss $\mathcal{L}_p$ for self-supervised learning using Equation (\ref{eq:self_loss}).
    \STATE Apply $k$-means clustering to partition $D^t$ into $K$ clusters.
    \STATE Compute centroids $\mathbf{c}_k^t$ for each cluster using Equation (\ref{eq:centclus}).
    \STATE Compare centroids using Equation (\ref{eq:cosine}) and assign pseudo-labels.
    \STATE Merge new data with memory $\mathcal{M}$ using Equation (\ref{eq:combined_data}).
\ENDFOR
\STATE \textbf{Meta Loop:} 
\STATE Compute entropy for each sample in memory $\mathcal{M}$ to prioritize low-entropy samples using Equation (\ref{eq:entro}).
\STATE Update memory buffer $\mathcal{M}$ with representative samples.
\STATE Train model on selected samples using cross-entropy loss (Equation:\ref{eq:celoss}).
\end{algorithmic}
\end{algorithm}

\section{Experiments}

\subsection{Dataset}
We evaluate the proposed approach on valence-arousal classification using EEG signals. For this, we use three publicly available datasets including AMIGOS \cite{Correa2017AMIGOSAD}, DEAP \cite{Koelstra2012DEAPAD} and PPB-EMO \cite{Li2022AMP}. In AMIGOS dataset, 14 channeled EEG signals are collected at sampling rate of 128 Hz from $40$ subjects who watched $20$ emotional videos during the process. The preprocessed version of the dataset is available at \footnote{https://www.eecs.qmul.ac.uk/mmv/datasets/amigos/}. Participants rated their emotional responses on a continuous scale from $1$ (low) to $9$ (high) for arousal, valence, and dominance. Similarly, the DEAP dataset consists of 32-channeled data from $32$ subjects, each participating in $40$ trials. The data are categorized into four classes based on the quadrants of valence and arousal. The PPB-Emo dataset comprises data from 40 participants engaged in driving tasks while experiencing various emotions. During the experiment, drivers watched emotion-evoking videos and then performed driving tasks reflecting those emotions. 32-channel EEG data was recorded at 250 Hz using the EnobioNE, a wireless EEG device. Additionally, participants provided self-reported ratings of valence, arousal, and dominance for each emotional state on a 9-point scale. Each dataset is segmented into a length of seven seconds with $50\%$ overlap between two consecutive segments.

\subsection{Experimental setup}
To adapt a pre-trained model from the source domain to an online continual target domain, we create two different simulations. In the first setting, we use PPB-EMO as the source domain and utilize DEAP in an OCL paradigm. The EEG model is first trained offline on the PPB-EMO dataset for four-class valence arousal classification. Then, the pre-trained model is adapted in the OCL environment where small batches of EEG data are streamed sequentially based on emotion elicitation videos per subject. This reflects class and subject-incremental settings, where each specific video represents a specific class and the model is agnostic to class IDs and subject IDs of the upcoming data. In the second experiment, a similar setting is used, but with the PPB-EMO and DEAP datasets as the source domain and AMIGOS as the target dataset. Since the AMIGOS dataset has only 14 EEG channels, we pre-train the model on both PPB-EMO and DEAP with only 14 channels.

\subsection{Baselines}
We compare our method with both supervised and self-supervised learning approaches. For the supervised learning baseline, EWC \cite{Kirkpatrick2016OvercomingCF}, SupCon \cite{khosla2020supervised}, CLUDA \cite{Ozyurt2022ContrastiveLF}, and AMBM \cite{Duan2024RetainAA} are considered. For self-supervised learning baselines, we use SimCLR \cite{Chen2020ASF}, and SCALE \cite{Yu2022SCALEOS}.

\begin{table}[t]
  \centering
    \caption{Accuracies $(\%)$ achieved on Test set on DEAP dataset. PPB-EMO dataset is used as source data for pre-training. Bold fonts show superior performance.}
  \label{tab:Deap_data}
  \begin{tabular}{cccc}
    \hline
    Baseline   & AdapAcc $(\%)$ & GenAcc $(\%)$ &  ForAcc $(\%)$ \\ \hline
    EWC        & $69.35 \pm 3.33$ & $29.99 \pm 1.02$ & $-41.09 \pm 3.56$ \\ 
    SupCon     & $61.53 \pm 1.69$ & $40.51 \pm 1.50$ & $-19.51 \pm 1.85$ \\ 
    CLUDA       & $60.02 \pm 1.34$ & $37.52 \pm 1.66$ & $-21.39 \pm 1.39$ \\
    AMBM       & $60.97 \pm 1.24$ & $35.05 \pm 0.86$ & $-28.31 \pm 1.69$ \\ 
    SimCLR     & $60.12 \pm 1.83$ & $37.85 \pm 1.51$ & $-21.12 \pm 1.43$ \\
    SCALE      & $58.26 \pm 1.90$ & $39.04 \pm 1.15$ & $-18.23 \pm 1.24$ \\ 
    SSOCL   &\bm{ $83.76 \pm 04.23$} & \bm{$71.78 \pm 5.69$} & \bm{$-15.05 \pm 3.89$} \\
    \hline
  \end{tabular}

\end{table}
\subsection{Model architecture and settings}

The feature extractor $f$ in this work is designed based on a one 1D Convolution layer, followed by three residual blocks ( two convolution layers per block with kernel sizes 15,21 and 43, a batch normalization and a max pooling layer (kernel size 4, stride 4). An attention layer is applied followed by three fully connected layers (1024, 512, 256 units) with ReLU activation, a projection head (128-embedding dimension), and a linear classifier $\phi$.
Adam optimizer is used with learning rate of $1e^{-4}$ and weight decay factor of $1e^{-4}$. Streamed data come into the batch size of $32$ segments. The memory buffer is fixed to $200$ samples and memory data batch size is set to $32$. $10$ steps of training are considered in meta-loop to train the model at each time step $t$ and $T=10$ is set for memory enhancement module. We apply a random affine transformation with up to 10 degrees of rotation and $1\%$ translation to samples for data augmentation in baseline methods that use contrastive loss.

\textbf{Evaluation Metrics:} We report Average Adaptation Accuracy (AdapAcc), Averaged Generalization Accuracy (GenAcc), and Average Forgetting  Accuracy (ForAcc) in our results. AdapAcc evaluates the performance on the current subject's test set immediately after training on that subject, calculated as $AdapAcc = \frac{1}{|\mathcal{S}|}\sum_{j=1}^{|\mathcal{S}|} a_j$, where \(a_j\) represents the accuracy on the test set for the subject \(j\). GenAcc is the average accuracy on the whole test dataset that is comprised of all subjects. ForAcc measures total forgetting and the ability of the model to retain knowledge from all seen subjects after training is completed for the current subject, expressed as $\frac{1}{{|\mathcal{S}|}-1} \sum_{j=1}^{{|\mathcal{S}|}-1} (a_j - m_j)$,  where \(m_j\) indicates the model's accuracy on the test set of the subject \(j\) at the end of training of $j+1$ and $a_j$ is adaptation accuracy of $j$. The last subject is omitted as it does not experience forgetting.

\subsection{Results}
\begin{table}[t]
  \centering
  \caption{Accuracies $(\%)$ achieved on Test set on AMIGOS dataset. PPB-EMO and DEAP datasets are used as source data. Bold fonts show superior performance.}
  \label{tab:amigos_data}
  \begin{tabular}{cccc}
    \hline

    Baseline   & AdapAcc $(\%)$ & GenAcc $(\%)$ &  ForAcc $(\%)$ \\ \hline
    EWC        & \bm{$72.32 \pm 1.97$} & $32.08 \pm 0.56$ & $-39.97 \pm 0.85$ \\ 
    SupCon     & $54.84 \pm 1.69$ & $37.24 \pm 1.19$ & $-17.61 \pm 1.83$ \\ 
    CLUDA     & $49.55 \pm 1.94$ & $34.37 \pm 1.45$ & $-15.19 \pm 1.88$ \\ 
    AMBM       & $61.54 \pm 1.99$ & $32.40 \pm 1.26$ & $-29.17 \pm 1.17$ \\ 
    SimCLR     & $49.78 \pm 1.76$ & $34.90 \pm 1.22$ & $-14.93 \pm 1.65$ \\      
    SCALE      & $47.05 \pm 1.76$ & $33.03 \pm 1.59$ & $-14.12 \pm 1.19$ \\ 
    SSOCL   & $62.75 \pm 6.60$ & \bm{$56.51 \pm 9.39$} & \bm{$-5.90 \pm 3.30$} \\ 
    \hline
  \end{tabular}
\end{table}
\begin{figure}
    \centering
    \includegraphics[height=0.50\linewidth]{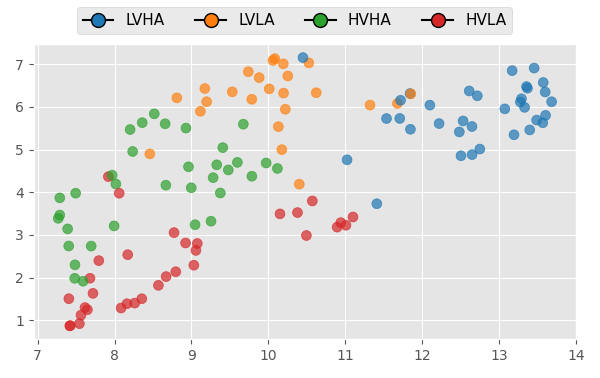}
    
    \caption{The memory buffer obtained using the SSOCL approach at the end of training on the DEAP dataset. The embeddings are generated using the model trained with SSOCL, and UMAP is applied to visualize these embeddings in a 2-dimensional plane. Colors represent the pseudo labels assigned to the stored samples.}
    \label{fig:mem_buff}
\end{figure}
Table \ref{tab:Deap_data} and \ref{tab:amigos_data} show the performance of the DEAP and AMIGOS datasets, respectively.  Mean and standard deviation are reported from five different runs, each with different subject sequences and initializations. First, we would like to highlight the performance of supervised learning approaches. SupCon is a supervised learning approach that uses supervised contrastive learning loss to optimize the model. Note that this approach utilizes data augmentation to generate positive and negative augmented examples, which are then used to compute the contrastive loss. However, data augmentation techniques may not be feasible for EEG data due to its low signal-to-noise ratio. DA can introduce extra noise and potentially alter the temporal information in the EEG signal, leading to suboptimal performance in this setting.
\begin{figure*}[!t] 
    \centering
    \subfloat[Subject: 1]{\includegraphics[width=0.30\textwidth]{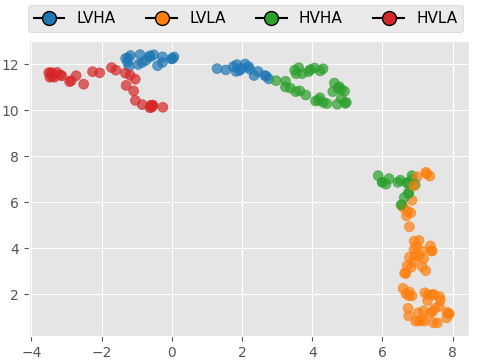}} \hfill
    \subfloat[Subject: 3]{\includegraphics[width=0.30\textwidth]{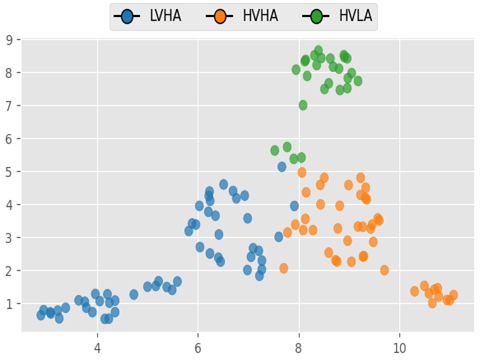}} \hfill
    \subfloat[Subject: 25]{\includegraphics[width=0.30\textwidth]{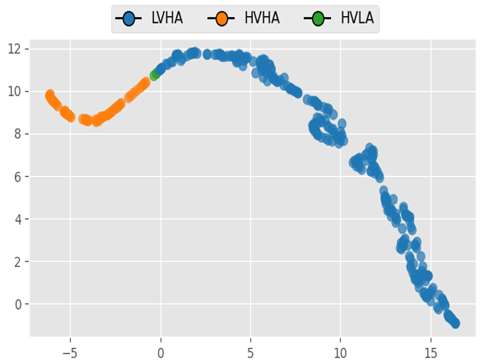}} \hfill
    \subfloat[Subject: 27]{\includegraphics[width=0.30\textwidth]{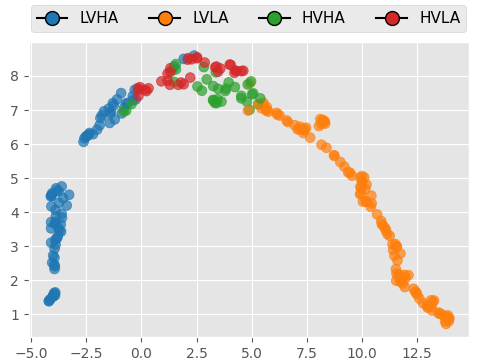}} \hfill
    \subfloat[Subject: 39]{\includegraphics[width=0.30\textwidth]{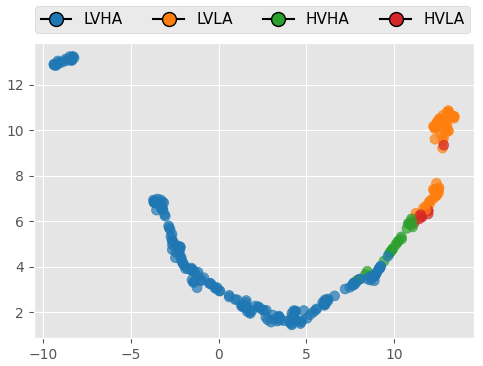}} \hfill
    \subfloat[Subject: 40]{\includegraphics[width=0.30\textwidth]{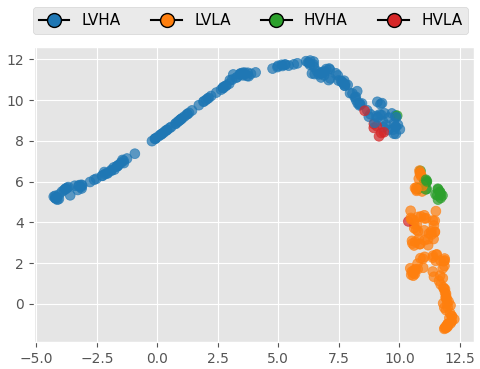}} \\
    
    \subfloat[Subject: 1]{\includegraphics[width=0.30\textwidth]{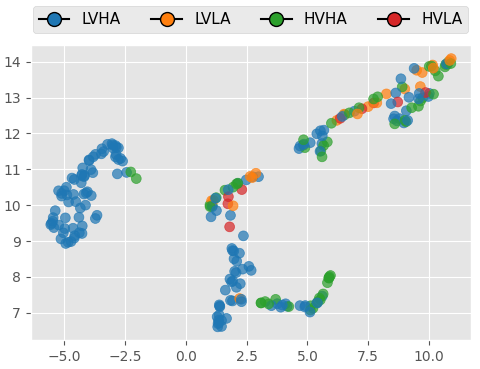}} \hfill
    \subfloat[Subject: 3]{\includegraphics[width=0.30\textwidth]{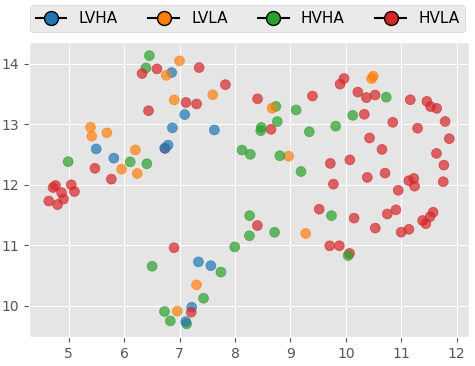}} \hfill
    \subfloat[Subject: 25]{\includegraphics[width=0.30\textwidth]{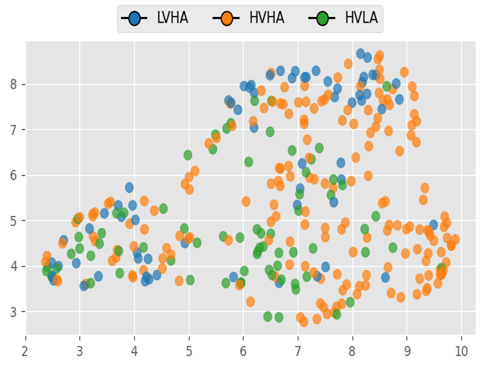}} \hfill
    \subfloat[Subject: 27]{\includegraphics[width=0.30\textwidth]{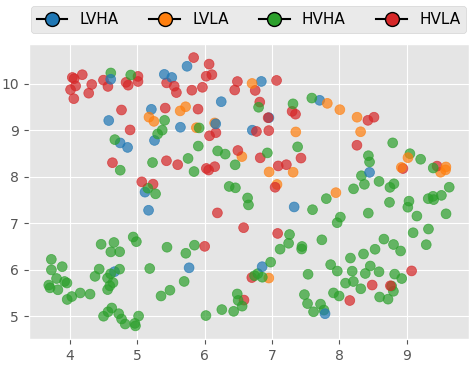}} \hfill
    \subfloat[Subject: 39]{\includegraphics[width=0.30\textwidth]{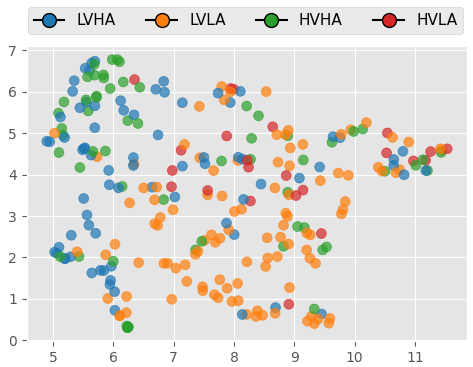}} \hfill
    \subfloat[Subject: 40]{\includegraphics[width=0.30\textwidth]{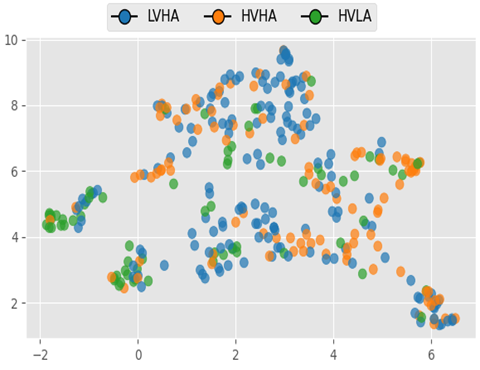}}
    
    \caption{Visualization of the embedding spaces learned by \textbf{SSOCL} (top rows: (a–f)) and \textbf{SCALE} (bottom rows: (g–l)). The feature extractor trained by each method on the AMIGOS dataset in OCL settings is used to compute embedding vectors for the test datasets of a few subjects. These embeddings are visualized with UMAP in a 2D plane. Each panel represents the embedding space for a particular subject, where each point corresponds to a test sample and the colors indicate the true labels.}
    \label{fig:scale_vs_SSOCL}
\end{figure*}
EWC and CLUDA are offline DA baselines. Since EWC incorporates the Fisher Information to constrain the loss function, it does not perform well in OCL settings. The Fisher Information is highly influenced by the quantity and quality of data \cite{Ahmad2023RobustFL}, which is limited in OCL settings due to the small number of samples available at each step $t$. This limitation prevents the model from capturing the important model parameters necessary for learning across subjects.

CLUDA is another offline DA approach that integrates data augmentation techniques into the framework of contrastive learning. As a result, CLUDA struggles with the dynamic and evolving nature of the data, leading to reduced effectiveness when handling subject shifts in EEG data. Furthermore, AMBM is a supervised learning approach specifically designed for continual learning with sequentially arriving EEG data. This method explicitly addresses inter-subject variability using supervised contrastive learning with data augmentation techniques. However, AMBM also does not perform well in the presence of both class-incremental and inter-subject variability. The tables show that all supervised learning baselines struggle to handle the heterogeneous nature of the continual learning setting in EEG data. This finding is consistent with recent research suggesting that SSL-based representational learning is more robust than traditional supervised approaches in OCL settings \cite{Liu2021SelfsupervisedLI,Fini2021SelfSupervisedMA}.

Now we turn our attention to the self-supervised baselines. SimCLR is an unsupervised learning approach designed for offline learning, which allows multiple passes over data. It shows poor performance in OCL settings where only a single pass is allowed. However, it still learns better-generalized representations compared to other methods including EWC, CLUDA, and AMBM. This is likely due to its ability to capture high-level features through contrastive learning, which helps in creating more robust representations that generalize well across different tasks. SCALE, on the other hand, is specifically designed for settings like ours and demonstrates better generalization compared to other baselines. Its ability to handle single-pass learning while maintaining a strong performance is a key advantage. However, SCALE is particularly prone to the inherent characteristics of EEG data, such as noise and variability, which can be amplified by its contrastive learning approach that heavily relies on data augmentation techniques. While data augmentation can help in some cases, it might not fully mitigate the challenges posed by the complex, noisy nature of EEG signals, potentially limiting the effectiveness of SCALE in certain scenarios. Finally, SSOCL outperforms all baselines across all performance metrics, demonstrating its ability to efficiently utilize the refined memory buffer samples. This demonstrates that SSOCL effectively creates a refined memory buffer using predictive loss, enabling robust generalization across all subjects.

Another interesting observation is that all alternatives show poor performance on AMIGOS as compared to the DEAP dataset. This reflects the importance of spatial feature diversity of EEG data. Since the DEAP dataset has 32 channels, they capture a broader and more detailed range of brain activity across different regions, allowing for a richer representation of the emotional states. In contrast, the AMIGOS dataset has fewer channels, which may limit the ability to capture complex spatial patterns and dynamics that are crucial for accurately identifying emotions. The reduced number of channels in AMIGOS could therefore lead to performance degradation, as less spatial information is available for analysis.

Figure~\ref{fig:mem_buff} shows the memory buffer obtained with the proposed approach at the end of the training on the DEAP dataset in the OCL setting. The same trained model is used to extract embeddings from the samples stored in the memory, which are visualized using the UMAP \footnote{https://umap-learn.readthedocs.io/en/latest/} dimensionality reduction approach. The visualization shows that the memory buffer effectively represents samples from all classes, with clear boundaries between them. The intra-class embeddings are well separated from each other, suggesting that diverse and generalized samples are stored in the memory buffer.

Moreover, Figure \ref{fig:scale_vs_SSOCL} shows the results of our final experiment, in which we compared the output of the feature extractor trained with SSOCL and SCALE. We selected random subjects, computed the embedding vectors of their test sets, and applied UMAP for visualization. The feature extractor trained with the SCALE approach struggles to separate inter-class clusters in the embedding space, resulting in significant overlap. This limitation could be due to the data augmentation strategy used for representation learning, which is unlikely to preserve temporal dependencies between samples. In contrast, the feature extractor trained with the SSOCL approach has a better organized embedding space, where inter-class embeddings form reasonably distinct clusters. While the cluster boundaries appear blurred or slightly overlapping, this is expected due to the inherent nature of EEG data, where clear boundaries are absent,and emotional states transition gradually from one to another.

\section{\textcolor{black}{Ablation Study}}
\subsection{Effect of Temperature $(T)$ on Memory Buffer}
\begin{figure*}[!t] 
    \centering
    \subfloat[$T = 1$]{\includegraphics[width=0.48\textwidth]{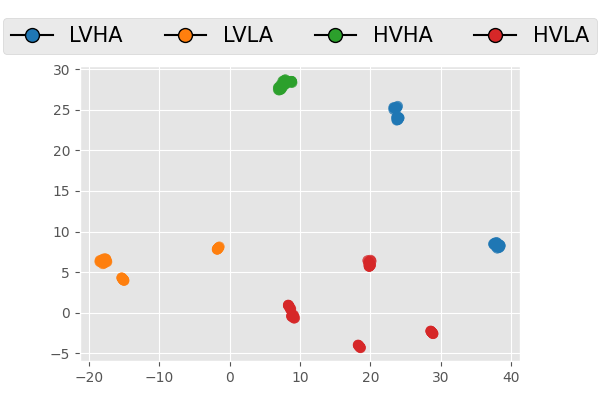}} \hfill
    \subfloat[$T = 10$]{\includegraphics[width=0.48\textwidth]{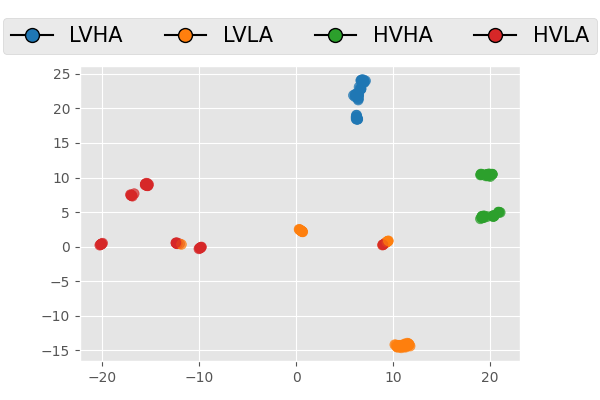}} \hfill
    \subfloat[$T = 100$]{\includegraphics[width=0.48\textwidth]{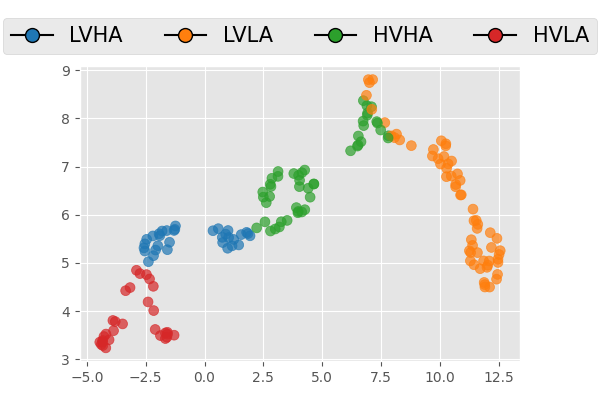}} \hfill
    \subfloat[$T = 1000$]{\includegraphics[width=0.48\textwidth]{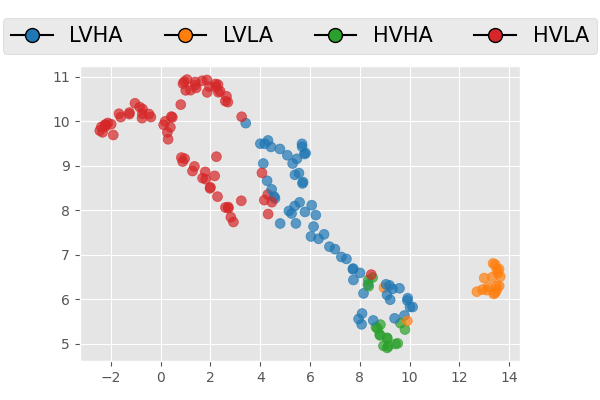}} \\
    
    \caption{UMAP visualizations of memory buffer embeddings for different temperature values: (a) \(T = 1\), (b) \(T = 10\), (c) \(T = 100\), and (d) \(T = 1000\). At \(T = 1\) and \(T = 10\), intra-class embeddings are highly clustered with minimal diversity. At \(T = 100\), clear and well-separated class boundaries emerge. At \(T = 1000\), the embeddings become blurred, with significant overlap between clusters, indicating a loss of distinct boundaries.}
    \label{fig:temp_effect}
\end{figure*}

This ablation study explores the impact of temperature \(T\) on the diversity and separability of embeddings within the memory buffer. As shown in Figure \ref{fig:temp_effect}, at lower temperatures (\(T = 1\) and \(T = 10\)), the embeddings are tightly clustered, exhibiting minimal intra-class variation, which limits diversity. A moderate temperature (\(T = 100\)) achieves an optimal balance, maintaining distinct class boundaries while ensuring sufficient intra-class variability. However, at very high temperatures (\(T = 1000\)), the embeddings become blurred and overlap, leading to a degradation in class separation and overall model performance.

\subsection{Impact of Self-Supervised and Memory Enhancement Modules}

\begin{table}[t]
  \centering
    \caption{SSOCL is compared with two ablated variants: one without the Self-Supervised Module (SSM), replaced by SimCLR-style contrastive learning; and another without the Memory Enhancement Module (MEnM), replaced by random memory sampling. 
}

  \label{tab:abalation_DEAP}
  \begin{tabular}{cccc}
    \hline

    Variant   & AdapAcc $(\%)$ & GenAcc $(\%)$ &  ForAcc $(\%)$ \\ \hline    
    SSOCL (w/o SSM) & $72.19$ & $68.18$ & $-13.26$ \\
    SSOCL (w/o MEnM) & $77.26$ & $64.85$ & $-14.63$ \\ 
    SSOCL   & $83.18$ & $72.96$ & $-9.92$ \\ 
    \hline
  \end{tabular}
\end{table}
Table~\ref{tab:abalation_DEAP} highlights the effectiveness of the proposed SSOCL framework by comparing it with two ablated variants. When the Self-Supervised Module (SSM) is removed and replaced with InfoNCE loss for contrastive learning \cite{chen2020simple}, the adaptation accuracy (AdapAcc) drops to $72.19\%$ and the generalization accuracy (GenAcc) to $68.18\%$, with a notable forgetting score (ForAcc) of $-13.26\%$. This degradation confirms that conventional contrastive learning with instance-level DA is less effective for EEG data. As discussed in Section~\ref{sec:sslm}, DA techniques can alter critical temporal dependencies in EEG signals, which are already affected by a low signal-to-noise ratio \cite{goldenholz2009mapping}. This results in noisy or less representative embeddings, reducing downstream performance.

Similarly, removing the Memory Enhancement Module (MEnM) and replacing it with random memory sampling yields $77.26\%$ AdapAcc and a sharper decline in GenAcc ($64.85\%$), along with increased forgetting ($-14.63\%$). This highlights the importance of entropy-based memory curation in preserving class boundaries and supporting continual learning. In contrast, the full SSOCL framework achieves the highest performance across all metrics, demonstrating that both self-supervised temporal modeling and entropy-aware memory design are crucial for effective adaptation and minimizing forgetting.

\section{Conclusion}
In this paper, we propose SSOCL, a self-supervised approach for OCL designed to address inter-subject variability without relying on subjective labels. SSOCL maintains a dynamic replay data buffer that evolves with the input data stream to effectively store generalized patterns. By incorporating a prediction network, SSOCL enables fast adaptation to the input data stream while preserving the temporal dependencies of EEG data. Moreover, it utilizes a robust cluster mapping mechanism to assign pseudo-labels to the input data stream. The proposed approach effectively selects diverse and representative samples and ensures that the memory buffer contains generalized and distinct examples for efficient replay. Experimental evaluations demonstrate the superior performance of SSOCL and highlight its robustness in maintaining a high-quality memory replay buffer.

\section{Funding}
This research was funded by the Asian Office of Aerospace Research and Development through Grant 23IOA087.

\bibliographystyle{elsarticle-harv}
\bibliography{manuscript.bib}

\begin{thebibliography}{45}
\expandafter\ifx\csname natexlab\endcsname\relax\def\natexlab#1{#1}\fi
\providecommand{\url}[1]{\texttt{#1}}
\providecommand{\href}[2]{#2}
\providecommand{\path}[1]{#1}
\providecommand{\DOIprefix}{doi:}
\providecommand{\ArXivprefix}{arXiv:}
\providecommand{\URLprefix}{URL: }
\providecommand{\Pubmedprefix}{pmid:}
\providecommand{\doi}[1]{\href{http://dx.doi.org/#1}{\path{#1}}}
\providecommand{\Pubmed}[1]{\href{pmid:#1}{\path{#1}}}
\providecommand{\bibinfo}[2]{#2}
\ifx\xfnm\relax \def\xfnm[#1]{\unskip,\space#1}\fi
\bibitem[{Ahmad et~al.(2023)Ahmad, Luo and Robles-Kelly}]{Ahmad2023RobustFL}
\bibinfo{author}{Ahmad, A.}, \bibinfo{author}{Luo, W.}, \bibinfo{author}{Robles-Kelly, A.}, \bibinfo{year}{2023}.
\newblock \bibinfo{title}{Robust federated learning under statistical heterogeneity via hessian-weighted aggregation}.
\newblock \bibinfo{journal}{Machine Learning} \bibinfo{volume}{112}, \bibinfo{pages}{633--654}.
\newblock \URLprefix \url{https://api.semanticscholar.org/CorpusID:255427209}.
\bibitem[{Aric{\'o} et~al.(2016a)Aric{\'o}, Borghini, di~Flumeri, Colosimo, Bonelli, Golfetti, Pozzi, Imbert, Granger, Benhac{\`e}ne and Babiloni}]{Aric2016AdaptiveAT}
\bibinfo{author}{Aric{\'o}, P.}, \bibinfo{author}{Borghini, G.}, \bibinfo{author}{di~Flumeri, G.}, \bibinfo{author}{Colosimo, A.}, \bibinfo{author}{Bonelli, S.}, \bibinfo{author}{Golfetti, A.}, \bibinfo{author}{Pozzi, S.}, \bibinfo{author}{Imbert, J.P.}, \bibinfo{author}{Granger, G.}, \bibinfo{author}{Benhac{\`e}ne, R.}, \bibinfo{author}{Babiloni, F.}, \bibinfo{year}{2016}a.
\newblock \bibinfo{title}{Adaptive automation triggered by eeg-based mental workload index: A passive brain-computer interface application in realistic air traffic control environment}.
\newblock \bibinfo{journal}{Frontiers in Human Neuroscience} \bibinfo{volume}{10}.
\newblock \URLprefix \url{https://api.semanticscholar.org/CorpusID:6717657}.
\bibitem[{Aric{\'o} et~al.(2016b)Aric{\'o}, Borghini, Flumeri, Colosimo, Pozzi and Babiloni}]{Aric2016APB}
\bibinfo{author}{Aric{\'o}, P.}, \bibinfo{author}{Borghini, G.}, \bibinfo{author}{Flumeri, G.D.}, \bibinfo{author}{Colosimo, A.}, \bibinfo{author}{Pozzi, S.}, \bibinfo{author}{Babiloni, F.}, \bibinfo{year}{2016}b.
\newblock \bibinfo{title}{A passive brain-computer interface application for the mental workload assessment on professional air traffic controllers during realistic air traffic control tasks.}
\newblock \bibinfo{journal}{Progress in brain research} \bibinfo{volume}{228}, \bibinfo{pages}{295--328}.
\newblock \URLprefix \url{https://api.semanticscholar.org/CorpusID:205167096}.
\bibitem[{Cao et~al.(2023)Cao, Yang, Zhang, Xi, Tang and Tian}]{Cao2023LURAO}
\bibinfo{author}{Cao, G.}, \bibinfo{author}{Yang, L.}, \bibinfo{author}{Zhang, Q.}, \bibinfo{author}{Xi, J.}, \bibinfo{author}{Tang, C.}, \bibinfo{author}{Tian, Y.}, \bibinfo{year}{2023}.
\newblock \bibinfo{title}{Lur: An online learning model for eeg emotion recognition}.
\newblock \bibinfo{journal}{2023 IEEE International Conference on Bioinformatics and Biomedicine (BIBM)} , \bibinfo{pages}{1038--1045}\URLprefix \url{https://api.semanticscholar.org/CorpusID:267043725}.
\bibitem[{Chen et~al.(2020a)Chen, Kornblith, Norouzi and Hinton}]{chen2020simple}
\bibinfo{author}{Chen, T.}, \bibinfo{author}{Kornblith, S.}, \bibinfo{author}{Norouzi, M.}, \bibinfo{author}{Hinton, G.}, \bibinfo{year}{2020}a.
\newblock \bibinfo{title}{A simple framework for contrastive learning of visual representations}, in: \bibinfo{booktitle}{Proceedings of the 37th International Conference on Machine Learning}, pp. \bibinfo{pages}{1597--1607}.
\bibitem[{Chen et~al.(2020b)Chen, Kornblith, Norouzi and Hinton}]{DBLP:conf/icml/ChenK0H20}
\bibinfo{author}{Chen, T.}, \bibinfo{author}{Kornblith, S.}, \bibinfo{author}{Norouzi, M.}, \bibinfo{author}{Hinton, G.E.}, \bibinfo{year}{2020}b.
\newblock \bibinfo{title}{A simple framework for contrastive learning of visual representations}, in: \bibinfo{booktitle}{ICML}, pp. \bibinfo{pages}{1597--1607}.
\newblock \URLprefix \url{http://proceedings.mlr.press/v119/chen20j.html}.
\bibitem[{Chen et~al.(2020c)Chen, Kornblith, Norouzi and Hinton}]{Chen2020ASF}
\bibinfo{author}{Chen, T.}, \bibinfo{author}{Kornblith, S.}, \bibinfo{author}{Norouzi, M.}, \bibinfo{author}{Hinton, G.E.}, \bibinfo{year}{2020}c.
\newblock \bibinfo{title}{A simple framework for contrastive learning of visual representations}.
\newblock \bibinfo{journal}{ArXiv} \bibinfo{volume}{abs/2002.05709}.
\newblock \URLprefix \url{https://api.semanticscholar.org/CorpusID:211096730}.
\bibitem[{Correa et~al.(2017)Correa, Abadi, Sebe and Patras}]{Correa2017AMIGOSAD}
\bibinfo{author}{Correa, J.A.M.}, \bibinfo{author}{Abadi, M.K.}, \bibinfo{author}{Sebe, N.}, \bibinfo{author}{Patras, I.}, \bibinfo{year}{2017}.
\newblock \bibinfo{title}{Amigos: A dataset for affect, personality and mood research on individuals and groups}.
\newblock \bibinfo{journal}{IEEE Transactions on Affective Computing} \bibinfo{volume}{12}, \bibinfo{pages}{479--493}.
\newblock \URLprefix \url{https://api.semanticscholar.org/CorpusID:8743034}.
\bibitem[{Deng et~al.(2018)Deng, Xu, Zhang, Fr{\"u}hholz and Schuller}]{Deng2018SemisupervisedAF}
\bibinfo{author}{Deng, J.}, \bibinfo{author}{Xu, X.}, \bibinfo{author}{Zhang, Z.}, \bibinfo{author}{Fr{\"u}hholz, S.}, \bibinfo{author}{Schuller, B.}, \bibinfo{year}{2018}.
\newblock \bibinfo{title}{Semisupervised autoencoders for speech emotion recognition}.
\newblock \bibinfo{journal}{IEEE/ACM Transactions on Audio, Speech, and Language Processing} \bibinfo{volume}{26}, \bibinfo{pages}{31--43}.
\newblock \URLprefix \url{https://api.semanticscholar.org/CorpusID:2212886}.
\bibitem[{Duan et~al.(2024a)Duan, Wang, Li, Doretto, Adjeroh, Yin and Tao}]{Duan2024OnlineCD}
\bibinfo{author}{Duan, T.}, \bibinfo{author}{Wang, Z.}, \bibinfo{author}{Li, F.}, \bibinfo{author}{Doretto, G.}, \bibinfo{author}{Adjeroh, D.}, \bibinfo{author}{Yin, Y.}, \bibinfo{author}{Tao, C.}, \bibinfo{year}{2024}a.
\newblock \bibinfo{title}{Online continual decoding of streaming eeg signal with a balanced and informative memory buffer.}
\newblock \bibinfo{journal}{Neural networks : the official journal of the International Neural Network Society} \bibinfo{volume}{176}, \bibinfo{pages}{106338}.
\newblock \URLprefix \url{https://api.semanticscholar.org/CorpusID:269403168}.
\bibitem[{Duan et~al.(2024b)Duan, Wang, Shen, Doretto, Adjeroh, Li and Tao}]{Duan2024RetainAA}
\bibinfo{author}{Duan, T.}, \bibinfo{author}{Wang, Z.}, \bibinfo{author}{Shen, L.}, \bibinfo{author}{Doretto, G.}, \bibinfo{author}{Adjeroh, D.}, \bibinfo{author}{Li, F.}, \bibinfo{author}{Tao, C.}, \bibinfo{year}{2024}b.
\newblock \bibinfo{title}{Retain and adapt: Online sequential eeg classification with subject shift}.
\newblock \bibinfo{journal}{IEEE Transactions on Artificial Intelligence} \URLprefix \url{https://api.semanticscholar.org/CorpusID:268984849}.
\bibitem[{Fini et~al.(2021)Fini, Costa, Alameda-Pineda, Ricci, Karteek and Mairal}]{Fini2021SelfSupervisedMA}
\bibinfo{author}{Fini, E.}, \bibinfo{author}{Costa, V.}, \bibinfo{author}{Alameda-Pineda, X.}, \bibinfo{author}{Ricci, E.}, \bibinfo{author}{Karteek, A.}, \bibinfo{author}{Mairal, J.}, \bibinfo{year}{2021}.
\newblock \bibinfo{title}{Self-supervised models are continual learners}.
\newblock \bibinfo{journal}{2022 IEEE/CVF Conference on Computer Vision and Pattern Recognition (CVPR)} , \bibinfo{pages}{9611--9620}\URLprefix \url{https://api.semanticscholar.org/CorpusID:244954199}.
\bibitem[{Goldenholz et~al.(2009)Goldenholz, Ahlfors, H{\"a}m{\"a}l{\"a}inen, Sharon, Ishitobi, Vaina and Stufflebeam}]{goldenholz2009mapping}
\bibinfo{author}{Goldenholz, D.M.}, \bibinfo{author}{Ahlfors, S.P.}, \bibinfo{author}{H{\"a}m{\"a}l{\"a}inen, M.S.}, \bibinfo{author}{Sharon, D.}, \bibinfo{author}{Ishitobi, M.}, \bibinfo{author}{Vaina, L.M.}, \bibinfo{author}{Stufflebeam, S.M.}, \bibinfo{year}{2009}.
\newblock \bibinfo{title}{Mapping the signal-to-noise-ratios of cortical sources in magnetoencephalography and electroencephalography}.
\newblock \bibinfo{journal}{Human brain mapping} \bibinfo{volume}{30}, \bibinfo{pages}{1077--1086}.
\bibitem[{Hafeez et~al.(2021)Hafeez, Saeed, Arsalan, Anwar, Ashraf and Alsubhi}]{Hafeez2021EEGIG}
\bibinfo{author}{Hafeez, T.}, \bibinfo{author}{Saeed, S.M.U.}, \bibinfo{author}{Arsalan, A.}, \bibinfo{author}{Anwar, S.M.}, \bibinfo{author}{Ashraf, M.U.}, \bibinfo{author}{Alsubhi, K.}, \bibinfo{year}{2021}.
\newblock \bibinfo{title}{Eeg in game user analysis: A framework for expertise classification during gameplay}.
\newblock \bibinfo{journal}{PLoS ONE} \bibinfo{volume}{16}.
\newblock \URLprefix \url{https://api.semanticscholar.org/CorpusID:231777428}.
\bibitem[{Hay(2012)}]{Hay2012EmotionRI}
\bibinfo{author}{Hay, M.O.}, \bibinfo{year}{2012}.
\newblock \bibinfo{title}{Emotion recognition in human-computer interaction}.
\newblock \URLprefix \url{https://api.semanticscholar.org/CorpusID:115023200}.
\bibitem[{Jia et~al.(2022)Jia, Ji, Zhou and Zhou}]{Jia2022HybridSN}
\bibinfo{author}{Jia, Z.}, \bibinfo{author}{Ji, J.}, \bibinfo{author}{Zhou, X.}, \bibinfo{author}{Zhou, Y.}, \bibinfo{year}{2022}.
\newblock \bibinfo{title}{Hybrid spiking neural network for sleep electroencephalogram signals}.
\newblock \bibinfo{journal}{Science China Information Sciences} \bibinfo{volume}{65}.
\newblock \URLprefix \url{https://api.semanticscholar.org/CorpusID:247591578}.
\bibitem[{Jiang et~al.(2024)Jiang, Lan and Lu}]{Jiang2024REmoNetRE}
\bibinfo{author}{Jiang, W.B.}, \bibinfo{author}{Lan, Y.T.}, \bibinfo{author}{Lu, B.L.}, \bibinfo{year}{2024}.
\newblock \bibinfo{title}{Remonet: Reducing emotional label noise via multi-regularized self-supervision}, in: \bibinfo{booktitle}{ACM Multimedia}.
\newblock \URLprefix \url{https://api.semanticscholar.org/CorpusID:273645536}.
\bibitem[{Jiang et~al.(2016)Jiang, Zheng, Tan, Tang and Zhou}]{Jiang2016VariationalDE}
\bibinfo{author}{Jiang, Z.}, \bibinfo{author}{Zheng, Y.}, \bibinfo{author}{Tan, H.}, \bibinfo{author}{Tang, B.}, \bibinfo{author}{Zhou, H.}, \bibinfo{year}{2016}.
\newblock \bibinfo{title}{Variational deep embedding: An unsupervised and generative approach to clustering}, in: \bibinfo{booktitle}{International Joint Conference on Artificial Intelligence}.
\newblock \URLprefix \url{https://api.semanticscholar.org/CorpusID:2546662}.
\bibitem[{Khare et~al.(2024)Khare, Blanes-Vidal, Nadimi and Acharya}]{khare2024emotion}
\bibinfo{author}{Khare, S.K.}, \bibinfo{author}{Blanes-Vidal, V.}, \bibinfo{author}{Nadimi, E.S.}, \bibinfo{author}{Acharya, U.R.}, \bibinfo{year}{2024}.
\newblock \bibinfo{title}{Emotion recognition and artificial intelligence: A systematic review (2014--2023) and research recommendations}.
\newblock \bibinfo{journal}{Information Fusion} \bibinfo{volume}{102}, \bibinfo{pages}{102019}.
\bibitem[{Khosla et~al.(2020)Khosla, Teterwak, Wang, Sarna, Tian, Isola, Maschinot, Liu and Krishnan}]{khosla2020supervised}
\bibinfo{author}{Khosla, P.}, \bibinfo{author}{Teterwak, P.}, \bibinfo{author}{Wang, C.}, \bibinfo{author}{Sarna, A.}, \bibinfo{author}{Tian, Y.}, \bibinfo{author}{Isola, P.}, \bibinfo{author}{Maschinot, A.}, \bibinfo{author}{Liu, C.}, \bibinfo{author}{Krishnan, D.}, \bibinfo{year}{2020}.
\newblock \bibinfo{title}{Supervised contrastive learning}.
\newblock \bibinfo{journal}{Advances in neural information processing systems} \bibinfo{volume}{33}, \bibinfo{pages}{18661--18673}.
\bibitem[{Kirkpatrick et~al.(2016)Kirkpatrick, Pascanu, Rabinowitz, Veness, Desjardins, Rusu, Milan, Quan, Ramalho, Grabska-Barwinska, Hassabis, Clopath, Kumaran and Hadsell}]{Kirkpatrick2016OvercomingCF}
\bibinfo{author}{Kirkpatrick, J.}, \bibinfo{author}{Pascanu, R.}, \bibinfo{author}{Rabinowitz, N.C.}, \bibinfo{author}{Veness, J.}, \bibinfo{author}{Desjardins, G.}, \bibinfo{author}{Rusu, A.A.}, \bibinfo{author}{Milan, K.}, \bibinfo{author}{Quan, J.}, \bibinfo{author}{Ramalho, T.}, \bibinfo{author}{Grabska-Barwinska, A.}, \bibinfo{author}{Hassabis, D.}, \bibinfo{author}{Clopath, C.}, \bibinfo{author}{Kumaran, D.}, \bibinfo{author}{Hadsell, R.}, \bibinfo{year}{2016}.
\newblock \bibinfo{title}{Overcoming catastrophic forgetting in neural networks}.
\newblock \bibinfo{journal}{Proceedings of the National Academy of Sciences} \bibinfo{volume}{114}, \bibinfo{pages}{3521 -- 3526}.
\newblock \URLprefix \url{https://api.semanticscholar.org/CorpusID:4704285}.
\bibitem[{Koelstra et~al.(2012)Koelstra, M{\"u}hl, Soleymani, Lee, Yazdani, Ebrahimi, Pun, Nijholt and Patras}]{Koelstra2012DEAPAD}
\bibinfo{author}{Koelstra, S.}, \bibinfo{author}{M{\"u}hl, C.}, \bibinfo{author}{Soleymani, M.}, \bibinfo{author}{Lee, J.S.}, \bibinfo{author}{Yazdani, A.}, \bibinfo{author}{Ebrahimi, T.}, \bibinfo{author}{Pun, T.}, \bibinfo{author}{Nijholt, A.}, \bibinfo{author}{Patras, I.}, \bibinfo{year}{2012}.
\newblock \bibinfo{title}{Deap: A database for emotion analysis ;using physiological signals}.
\newblock \bibinfo{journal}{IEEE Transactions on Affective Computing} \bibinfo{volume}{3}, \bibinfo{pages}{18--31}.
\newblock \URLprefix \url{https://api.semanticscholar.org/CorpusID:206597685}.
\bibitem[{Li et~al.(2020a)Li, Qiu, Du, Wang and He}]{Li2020DomainAF}
\bibinfo{author}{Li, J.}, \bibinfo{author}{Qiu, S.}, \bibinfo{author}{Du, C.}, \bibinfo{author}{Wang, Y.}, \bibinfo{author}{He, H.}, \bibinfo{year}{2020}a.
\newblock \bibinfo{title}{Domain adaptation for eeg emotion recognition based on latent representation similarity}.
\newblock \bibinfo{journal}{IEEE Transactions on Cognitive and Developmental Systems} \bibinfo{volume}{12}, \bibinfo{pages}{344--353}.
\newblock \URLprefix \url{https://api.semanticscholar.org/CorpusID:208091556}.
\bibitem[{Li et~al.(2020b)Li, Qiu, Shen, Liu and He}]{Li2020MultisourceTL}
\bibinfo{author}{Li, J.}, \bibinfo{author}{Qiu, S.}, \bibinfo{author}{Shen, Y.}, \bibinfo{author}{Liu, C.L.}, \bibinfo{author}{He, H.}, \bibinfo{year}{2020}b.
\newblock \bibinfo{title}{Multisource transfer learning for cross-subject eeg emotion recognition}.
\newblock \bibinfo{journal}{IEEE Transactions on Cybernetics} \bibinfo{volume}{50}, \bibinfo{pages}{3281--3293}.
\newblock \URLprefix \url{https://api.semanticscholar.org/CorpusID:89620182}.
\bibitem[{Li et~al.(2022)Li, Tan, Xing, Li, Li, Zeng, Wang, Zhang, Su, Pi, Guo and Cao}]{Li2022AMP}
\bibinfo{author}{Li, W.}, \bibinfo{author}{Tan, R.}, \bibinfo{author}{Xing, Y.}, \bibinfo{author}{Li, G.}, \bibinfo{author}{Li, S.}, \bibinfo{author}{Zeng, G.}, \bibinfo{author}{Wang, P.}, \bibinfo{author}{Zhang, B.}, \bibinfo{author}{Su, X.}, \bibinfo{author}{Pi, D.}, \bibinfo{author}{Guo, G.}, \bibinfo{author}{Cao, D.}, \bibinfo{year}{2022}.
\newblock \bibinfo{title}{A multimodal psychological, physiological and behavioural dataset for human emotions in driving tasks}.
\newblock \bibinfo{journal}{Scientific Data} \bibinfo{volume}{9}.
\newblock \URLprefix \url{https://api.semanticscholar.org/CorpusID:251369850}.
\bibitem[{Li et~al.(2020c)Li, Fu, Li, Shi and Zheng}]{Li2020ANT}
\bibinfo{author}{Li, Y.}, \bibinfo{author}{Fu, B.}, \bibinfo{author}{Li, F.}, \bibinfo{author}{Shi, G.}, \bibinfo{author}{Zheng, W.}, \bibinfo{year}{2020}c.
\newblock \bibinfo{title}{A novel transferability attention neural network model for eeg emotion recognition}.
\newblock \bibinfo{journal}{ArXiv} \bibinfo{volume}{abs/2009.09585}.
\newblock \URLprefix \url{https://api.semanticscholar.org/CorpusID:221819383}.
\bibitem[{Liang et~al.(2018)Liang, Li and Srikant}]{liang2018enhancing}
\bibinfo{author}{Liang, S.}, \bibinfo{author}{Li, Y.}, \bibinfo{author}{Srikant, R.}, \bibinfo{year}{2018}.
\newblock \bibinfo{title}{Enhancing the reliability of out-of-distribution image detection in neural networks}, in: \bibinfo{booktitle}{International Conference on Learning Representations}.
\newblock \URLprefix \url{https://openreview.net/forum?id=H1VGkIxRZ}.
\bibitem[{Lim(2021)}]{lim2021cognitive}
\bibinfo{author}{Lim, Y.X.}, \bibinfo{year}{2021}.
\newblock \bibinfo{title}{Cognitive Human-Machine Interfaces and Interactions for Avionics Systems}.
\newblock \bibinfo{type}{Phd thesis}. RMIT University.
\newblock \URLprefix \url{https://doi.org/10.25439/rmt.27601791}, \DOIprefix\doi{10.25439/rmt.27601791}.
\bibitem[{Liu et~al.(2024)Liu, qiu Zhou, Zhu, Zhai, Jia and Liu}]{Liu2024VBHGNNVB}
\bibinfo{author}{Liu, C.}, \bibinfo{author}{qiu Zhou, X.}, \bibinfo{author}{Zhu, Z.}, \bibinfo{author}{Zhai, L.}, \bibinfo{author}{Jia, Z.}, \bibinfo{author}{Liu, Y.}, \bibinfo{year}{2024}.
\newblock \bibinfo{title}{Vbh-gnn: Variational bayesian heterogeneous graph neural networks for cross-subject emotion recognition}, in: \bibinfo{booktitle}{International Conference on Learning Representations}.
\newblock \URLprefix \url{https://api.semanticscholar.org/CorpusID:271745983}.
\bibitem[{Liu et~al.(2021)Liu, HaoChen, Gaidon and Ma}]{Liu2021SelfsupervisedLI}
\bibinfo{author}{Liu, H.}, \bibinfo{author}{HaoChen, J.Z.}, \bibinfo{author}{Gaidon, A.}, \bibinfo{author}{Ma, T.}, \bibinfo{year}{2021}.
\newblock \bibinfo{title}{Self-supervised learning is more robust to dataset imbalance}.
\newblock \bibinfo{journal}{ArXiv} \bibinfo{volume}{abs/2110.05025}.
\newblock \URLprefix \url{https://api.semanticscholar.org/CorpusID:238583191}.
\bibitem[{Mou et~al.(2023)Mou, Zhao, Zhou, Nakisa, Rastgoo, Ma, Huang, Yin, Jain and Gao}]{Mou2023DriverER}
\bibinfo{author}{Mou, L.}, \bibinfo{author}{Zhao, Y.}, \bibinfo{author}{Zhou, C.}, \bibinfo{author}{Nakisa, B.}, \bibinfo{author}{Rastgoo, M.N.}, \bibinfo{author}{Ma, L.}, \bibinfo{author}{Huang, T.}, \bibinfo{author}{Yin, B.}, \bibinfo{author}{Jain, R.}, \bibinfo{author}{Gao, W.}, \bibinfo{year}{2023}.
\newblock \bibinfo{title}{Driver emotion recognition with a hybrid attentional multimodal fusion framework}.
\newblock \bibinfo{journal}{IEEE Transactions on Affective Computing} \bibinfo{volume}{14}, \bibinfo{pages}{2970--2981}.
\newblock \URLprefix \url{https://api.semanticscholar.org/CorpusID:257262644}.
\bibitem[{Nakisa et~al.(2018a)Nakisa, Rastgoo, Rakotonirainy, Maire and Chandran}]{Nakisa2018LongST}
\bibinfo{author}{Nakisa, B.}, \bibinfo{author}{Rastgoo, M.N.}, \bibinfo{author}{Rakotonirainy, A.}, \bibinfo{author}{Maire, F.}, \bibinfo{author}{Chandran, V.}, \bibinfo{year}{2018}a.
\newblock \bibinfo{title}{Long short term memory hyperparameter optimization for a neural network based emotion recognition framework}.
\newblock \bibinfo{journal}{IEEE Access} \bibinfo{volume}{6}, \bibinfo{pages}{49325--49338}.
\newblock \URLprefix \url{https://api.semanticscholar.org/CorpusID:52896590}.
\bibitem[{Nakisa et~al.(2018b)Nakisa, Rastgoo, Tjondronegoro and Chandran}]{Nakisa2018EvolutionaryCA}
\bibinfo{author}{Nakisa, B.}, \bibinfo{author}{Rastgoo, M.N.}, \bibinfo{author}{Tjondronegoro, D.}, \bibinfo{author}{Chandran, V.}, \bibinfo{year}{2018}b.
\newblock \bibinfo{title}{Evolutionary computation algorithms for feature selection of eeg-based emotion recognition using mobile sensors}.
\newblock \bibinfo{journal}{Expert Syst. Appl.} \bibinfo{volume}{93}, \bibinfo{pages}{143--155}.
\newblock \URLprefix \url{https://api.semanticscholar.org/CorpusID:3203588}.
\bibitem[{Ozyurt et~al.(2022)Ozyurt, Feuerriegel and Zhang}]{Ozyurt2022ContrastiveLF}
\bibinfo{author}{Ozyurt, Y.}, \bibinfo{author}{Feuerriegel, S.}, \bibinfo{author}{Zhang, C.}, \bibinfo{year}{2022}.
\newblock \bibinfo{title}{Contrastive learning for unsupervised domain adaptation of time series}.
\newblock \bibinfo{journal}{ArXiv} \bibinfo{volume}{abs/2206.06243}.
\newblock \URLprefix \url{https://api.semanticscholar.org/CorpusID:249625545}.
\bibitem[{Rastgoo et~al.(2019)Rastgoo, Nakisa, Maire, Rakotonirainy and Chandran}]{Rastgoo2019AutomaticDS}
\bibinfo{author}{Rastgoo, M.N.}, \bibinfo{author}{Nakisa, B.}, \bibinfo{author}{Maire, F.}, \bibinfo{author}{Rakotonirainy, A.}, \bibinfo{author}{Chandran, V.}, \bibinfo{year}{2019}.
\newblock \bibinfo{title}{Automatic driver stress level classification using multimodal deep learning}.
\newblock \bibinfo{journal}{Expert Syst. Appl.} \bibinfo{volume}{138}.
\newblock \URLprefix \url{https://api.semanticscholar.org/CorpusID:198344222}.
\bibitem[{Sarkar and Etemad(2020)}]{Sarkar2020SelfSupervisedER}
\bibinfo{author}{Sarkar, P.}, \bibinfo{author}{Etemad, A.}, \bibinfo{year}{2020}.
\newblock \bibinfo{title}{Self-supervised ecg representation learning for emotion recognition}.
\newblock \bibinfo{journal}{IEEE Transactions on Affective Computing} \bibinfo{volume}{13}, \bibinfo{pages}{1541--1554}.
\newblock \URLprefix \url{https://api.semanticscholar.org/CorpusID:211069320}.
\bibitem[{Tiezzi et~al.(2022)Tiezzi, Marullo, Faggi, Meloni, Betti and Melacci}]{Tiezzi2022StochasticCO}
\bibinfo{author}{Tiezzi, M.}, \bibinfo{author}{Marullo, S.}, \bibinfo{author}{Faggi, L.}, \bibinfo{author}{Meloni, E.}, \bibinfo{author}{Betti, A.}, \bibinfo{author}{Melacci, S.}, \bibinfo{year}{2022}.
\newblock \bibinfo{title}{Stochastic coherence over attention trajectory for continuous learning in video streams}.
\newblock \bibinfo{journal}{ArXiv} \bibinfo{volume}{abs/2204.12193}.
\newblock \URLprefix \url{https://api.semanticscholar.org/CorpusID:248392341}.
\bibitem[{Vazquez-Rodriguez et~al.(2022)Vazquez-Rodriguez, Lefebvre, Cumin and Crowley}]{VazquezRodriguez2022TransformerBasedSL}
\bibinfo{author}{Vazquez-Rodriguez, J.}, \bibinfo{author}{Lefebvre, G.}, \bibinfo{author}{Cumin, J.}, \bibinfo{author}{Crowley, J.L.}, \bibinfo{year}{2022}.
\newblock \bibinfo{title}{Transformer-based self-supervised learning for emotion recognition}.
\newblock \bibinfo{journal}{2022 26th International Conference on Pattern Recognition (ICPR)} , \bibinfo{pages}{2605--2612}\URLprefix \url{https://api.semanticscholar.org/CorpusID:248084923}.
\bibitem[{Wang et~al.(2023)Wang, Ma, Cammon, Fang, Gao and Zhang}]{Wang2023SelfSupervisedEE}
\bibinfo{author}{Wang, X.}, \bibinfo{author}{Ma, Y.}, \bibinfo{author}{Cammon, J.}, \bibinfo{author}{Fang, F.}, \bibinfo{author}{Gao, Y.}, \bibinfo{author}{Zhang, Y.}, \bibinfo{year}{2023}.
\newblock \bibinfo{title}{Self-supervised eeg emotion recognition models based on cnn}.
\newblock \bibinfo{journal}{IEEE Transactions on Neural Systems and Rehabilitation Engineering} \bibinfo{volume}{31}, \bibinfo{pages}{1952--1962}.
\newblock \URLprefix \url{https://api.semanticscholar.org/CorpusID:257891756}.
\bibitem[{Wang et~al.(2022)Wang, Shen, Fang, Suo, Duan and Gao}]{Wang2022ImprovingTC}
\bibinfo{author}{Wang, Z.}, \bibinfo{author}{Shen, L.}, \bibinfo{author}{Fang, L.}, \bibinfo{author}{Suo, Q.}, \bibinfo{author}{Duan, T.}, \bibinfo{author}{Gao, M.}, \bibinfo{year}{2022}.
\newblock \bibinfo{title}{Improving task-free continual learning by distributionally robust memory evolution}, in: \bibinfo{booktitle}{International Conference on Machine Learning}.
\newblock \URLprefix \url{https://api.semanticscholar.org/CorpusID:250360832}.
\bibitem[{Wu et~al.(2023)Wu, Daoudi and Amad}]{Wu2023TransformerBasedSM}
\bibinfo{author}{Wu, Y.}, \bibinfo{author}{Daoudi, M.}, \bibinfo{author}{Amad, A.}, \bibinfo{year}{2023}.
\newblock \bibinfo{title}{Transformer-based self-supervised multimodal representation learning for wearable emotion recognition}.
\newblock \bibinfo{journal}{IEEE Transactions on Affective Computing} \bibinfo{volume}{15}, \bibinfo{pages}{157--172}.
\newblock \URLprefix \url{https://api.semanticscholar.org/CorpusID:257901125}.
\bibitem[{Yu et~al.(2022)Yu, Guo, Gao and Simunic}]{Yu2022SCALEOS}
\bibinfo{author}{Yu, X.}, \bibinfo{author}{Guo, Y.}, \bibinfo{author}{Gao, S.}, \bibinfo{author}{Simunic, T.}, \bibinfo{year}{2022}.
\newblock \bibinfo{title}{Scale: Online self-supervised lifelong learning without prior knowledge}.
\newblock \bibinfo{journal}{2023 IEEE/CVF Conference on Computer Vision and Pattern Recognition Workshops (CVPRW)} , \bibinfo{pages}{2484--2495}\URLprefix \url{https://api.semanticscholar.org/CorpusID:251765472}.
\bibitem[{Zhang et~al.(2018)Zhang, Traue and Hazer-Rau}]{Zhang2018IndividualER}
\bibinfo{author}{Zhang, L.}, \bibinfo{author}{Traue, H.C.}, \bibinfo{author}{Hazer-Rau, D.}, \bibinfo{year}{2018}.
\newblock \bibinfo{title}{Individual emotion recognition and subgroup analysis from psychophysiological signals}.
\newblock \bibinfo{journal}{Electrical Engineering eJournal} .
\bibitem[{Zhang et~al.(2021)Zhang, hua Zhong and Liu}]{Zhang2021GANSERAS}
\bibinfo{author}{Zhang, Z.}, \bibinfo{author}{hua Zhong, S.}, \bibinfo{author}{Liu, Y.}, \bibinfo{year}{2021}.
\newblock \bibinfo{title}{Ganser: A self-supervised data augmentation framework for eeg-based emotion recognition}.
\newblock \bibinfo{journal}{IEEE Transactions on Affective Computing} \bibinfo{volume}{14}, \bibinfo{pages}{2048--2063}.
\newblock \URLprefix \url{https://api.semanticscholar.org/CorpusID:237431425}.
\bibitem[{Zhao et~al.(2021)Zhao, Yan and Lu}]{Zhao2021PlugandPlayDA}
\bibinfo{author}{Zhao, L.M.}, \bibinfo{author}{Yan, X.}, \bibinfo{author}{Lu, B.L.}, \bibinfo{year}{2021}.
\newblock \bibinfo{title}{Plug-and-play domain adaptation for cross-subject eeg-based emotion recognition}, in: \bibinfo{booktitle}{AAAI Conference on Artificial Intelligence}.
\newblock \URLprefix \url{https://api.semanticscholar.org/CorpusID:235306399}.

\end{thebibliography}
\end{document}